\documentclass[runningheads,a4paper]{llncs}
\usepackage[T1]{fontenc}
\usepackage{graphicx}
\usepackage{tabularx,booktabs}
\usepackage{hyperref}
\usepackage{cite}
\usepackage{amsmath,amssymb,amsfonts}
\usepackage{algorithmic}
\usepackage{multirow}
\usepackage{subcaption}
\usepackage{float}
\usepackage{authblk}
\usepackage{amssymb}
\usepackage{url}
\newcommand{\keywords}[1]{\par\addvspace\baselineskip
\noindent\keywordname\enspace\ignorespaces#1}

\begin{document}

\mainmatter  

\title{{MambaU-Lite: A Lightweight Model based on Mamba and Integrated Channel-Spatial Attention for Skin Lesion Segmentation}}
\titlerunning{MambaU-Lite}

%
%
\author{Thi-Nhu-Quynh Nguyen\and
Quang-Huy Ho \and Duy-Thai Nguyen \and Hoang-Minh-Quang Le \and Van-Truong Pham \and Thi-Thao Tran\inst{*}}
\authorrunning{ Preprint}

\institute{Department of Automation Engineering\\ 
School of Electrical and Electronic Engineering\\
Hanoi University of Science and Technology, Vietnam\\
$^*$Email: \textit{thao.tranthi@hust.edu.vn} 
}


%
%

\toctitle{MambaU-Lite Model Preprint}
\tocauthor{Thi-Nhu-Quynh Nguyen}
\maketitle
\begin{abstract}
Early detection of skin abnormalities plays a crucial role in diagnosing and treating skin cancer. Segmentation of affected skin regions using AI-powered devices is relatively common and supports the diagnostic process. However, achieving high performance remains a significant challenge due to the need for high-resolution images and the often unclear boundaries of individual lesions. At the same time, medical devices require segmentation models to have a small memory footprint and low computational cost. Based on these requirements, we introduce a novel lightweight model called MambaU-Lite, which combines the strengths of Mamba and CNN architectures, featuring just over 400K parameters and a computational cost of more than 1G flops. To enhance both global context and local feature extraction, we propose the P-Mamba block, a novel component that incorporates VSS blocks alongside multiple pooling layers, enabling the model to effectively learn multi-scale features and enhance segmentation performance. We evaluate the model's performance on two skin datasets, ISIC2018 and PH2, yielding promising results. Our source code is publicly available at: \href{https://github.com/nqnguyen812/MambaU-Lite}{https://github.com/nqnguyen812/MambaU-Lite}. \\
\end{abstract}
\section{Introduction}

The segmentation of skin lesions plays an important role in computer-aided diagnostic systems for skin cancer.  However, before automated technology made its step into this medical area, the manual method of segmentation was thought to be tedious and inaccurate, which is unreliable and costly overall. Fortunately, with the advances of deep learning, especially the U-Net\cite{ronneberger2015u} and variants \cite{valanarasu2022unext},\cite{oktay2018attention},\cite{pham2021ear}, various attempts to implement those into the segmentation tasks have been done with the aim of eliminating human error as well as increasing speed. 


On the other spectrum of Machine Learning and Neural Networks, in 2017, a new model called Transformer\cite{vaswani2017attention} with the core mechanism "attention" made a revolutionary breakthrough with how impressively the model dealt with NLP tasks. Ideally, to bridge the gaps between Transformer\cite{vaswani2017attention} in NLP tasks and some prior models in Computer Vision tasks, Dosovitskiy \textit{et al.} had proposed Vision Transformer\cite{dosovitskiy2020image}, including a component called "ViT", establishing a new era for various Transformer-based image processing models. Extending this idea into segmentation tasks, TransUNet\cite{chen2021transunet} integrated both the U-Net structure and the powerful ViT block. UCTransnet\cite{wang2022uctransnet}, Swin-Unet\cite{cao2022swin}, and various later models employed similar combinations with various adjustments and yielded relatively successful results. However, there lies a problem with Transformer and the attention mechanism\cite{vaswani2017attention}, that is the computational complexity of the mechanism scales quadratically with the sequence length, making the inference speed non-ideal in some cases. This applies to segmentation and image processing in general as well, where high-resolution images when flattened could result in an extremely long sequence.

Recently in 2024, with a different approach,  Gu and Dao proposed S6 model or so called Mamba\cite{gu2023mamba}, which improves the performance of the casual structured state spaces models (S4) by implementing the selection mechanism and the hardware-aware algorithm. Most importantly, this model scales linearly while yielding promising and competitive results compared to Transformer-based models. Making use of Mamba in image processing, Vision Mamba or Vim\cite{zhu2024vision} employed a bidirectional SSM mechanism for selectively capturing the global context of the image. Furthermore, VMamba\cite{liu2024vmambavisualstatespace} with the VSS block, built upon the 2D-selective-scan (SS2D) mechanism, was proposed later that year, allowing the model to learn the image from four directions, making the Mamba mechanism more compatible with image processing. Our hybrid model MambaU-Lite, inheriting the power of the VSS block, combined with the elegant design of U-Lite model\cite{dinh20231m}, has produced potentially good results while operating on only over 400K parameters.  
The following are the main contributions of our research:
\begin{itemize}
    \item We proposed a lightweight model, namely MambaU-Lite, a hybrid segmentation model integrating the uses of both Mamba and CNN, harnessing the best of all and levitating the performance while maintaining reasonable computation cost.
    \item A novel sub-structure called P-Mamba was established and implemented to efficiently learn features of different levels.
    \item The MambaU-Lite model was evaluated on two well-known skin lesion segmentation datasets, ISIC 2018 and PH2, producing promising results regarding the model being a lightweight one.
\end{itemize}
\section{Related Work}
\textbf{Visual State Space Model} \cite{liu2024vmambavisualstatespace}. Inspired by the Mamba \cite{gu2023mamba}, which successfully applied the State Space Model (SSM) from Control Theory to Natural Language Processing (NLP), Vision Mamba \cite{liu2024vmambavisualstatespace} was introduced as a novel approach to efficiently support visual representation by integrating SSM-based blocks. Additionally, this model not only facilitates the extraction of global features but also minimizes computational costs and time consumption. As a result, the application of SSM in vision-related tasks is becoming a trend \cite{rahman2024mamba}, and medical segmentation is no exception \cite{heidari2024computation}.

\textbf{U-Net architecture} \cite{ronneberger2015u}. U-Net, first introduced by Ronneberger \textit{et al.} in 2015, has laid the foundation for numerous medical image segmentation models. Featuring a straightforward architecture that follows a symmetric encoder-decoder pattern with skip connections, U-Net effectively addresses the challenge of limited labeled data and outperforms previous segmentation models in terms of efficiency. Subsequent improvements to U-Net, such as Attention U-Net \cite{oktay2018attention} have further affirmed this architecture's superiority in image segmentation.


\section{The Proposed Model}\label{The Proposed Model}
\begin{figure}[!h]
\centering
\includegraphics[width=\linewidth]{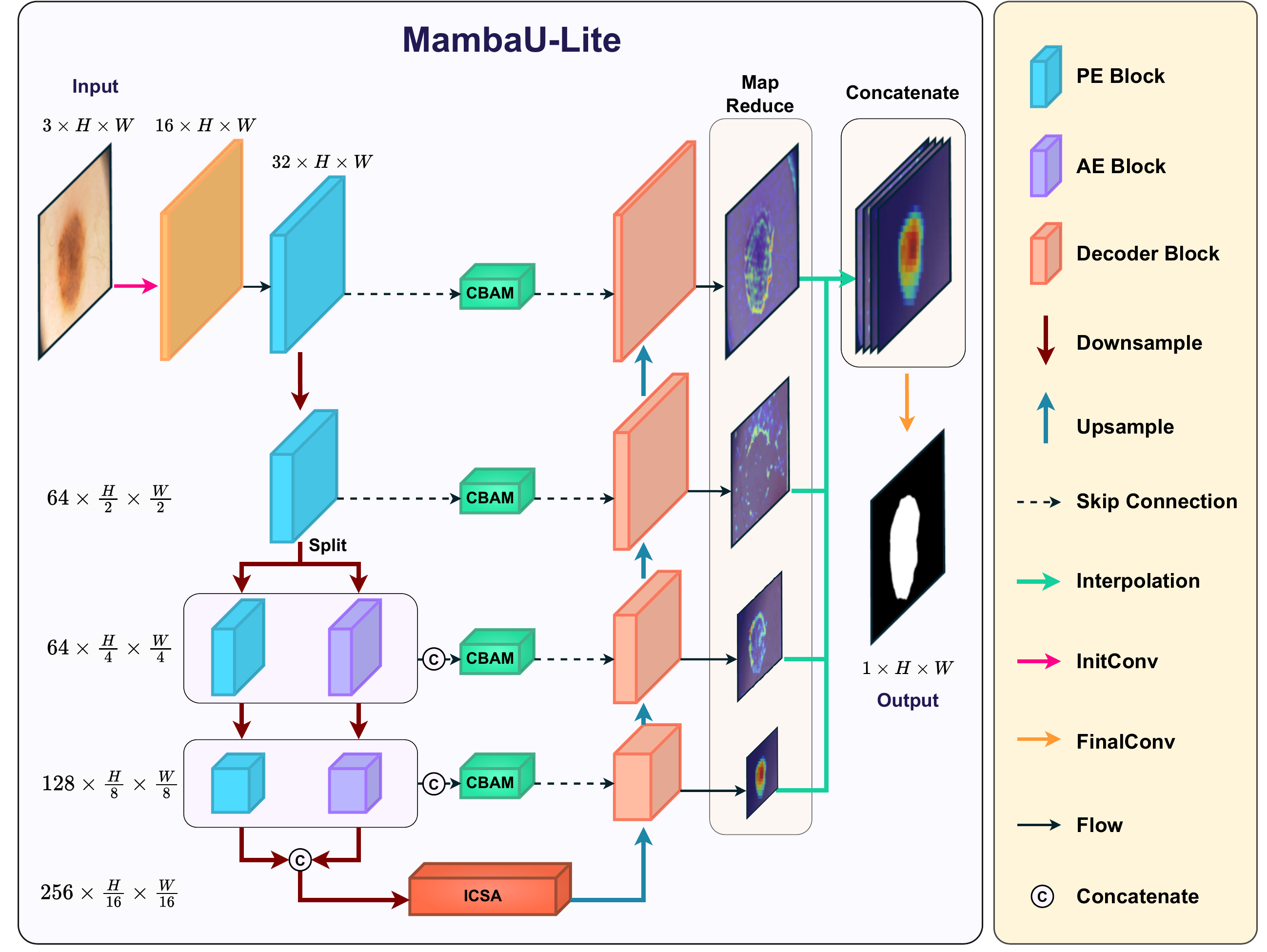}
\caption{The architecture of the proposed MambaU-Lite model}
\label{fig:model}
\end{figure}

In this section, the architecture of the proposed MambaU-Lite model is fully detailed and demonstrated in Fig.\ref{fig:model}. The model contains three fundamental sub-structures: Encoders, Bottleneck, and Decoders, together forming a U-shape combination similar to that of the classical U-Net \cite{ronneberger2015u}. Additionally, the model goes through four processing stages with four CBAM \cite{woo2018cbam} blocks in the Skip-connection assisting the Decoders with rich spatial information from the Encoders. 

Initially, the input image is passed through an InitConv layer to adjust the number of channels to 16, resulting in a feature map of size \( 16 \times H \times W \). The image then undergoes a sequence of Encoder layers. Specifically, in the proposed architecture, we use the first two P-Mamba Encoder blocks (PE Blocks). After these two blocks together with max-pooling layers to reduce the spatial dimensions after each Encoder, the feature map sizes are \( 32 \times \frac{H}{2} \times \frac{W}{2} \) and \( 64 \times \frac{H}{4} \times \frac{W}{4} \), respectively. For the next two Encoder layers, the input is split into two parts, effectively reducing the number of channels by half, and is then processed through the PE Block and the Axial Encoder Block (AE Block). The sizes of the feature maps after passing through both the PE and AE Blocks and max-pooling layers are identical, with dimensions \( 64 \times \frac{H}{8} \times \frac{W}{8} \) and \( 128 \times \frac{H}{16} \times \frac{W}{16} \), respectively. The outputs of the final two blocks are concatenated, resulting in a feature map of size \( 256 \times \frac{H}{16} \times \frac{W}{16} \) and fed to a bottleneck and then combined with skip connections, is passed through the Decoder layers. After passing through all the Decoders and upsampling layers, the output from each decoder block is interpolated back to the original input size. These outputs are subsequently concatenated and processed through a FinalConv layer to produce the predicted mask for the input image.

\subsection{The proposed PMamba Block}\label{The proposed PMamba Block}
\vspace{-3pt}
\begin{figure}[!h]
\centering
\includegraphics[width=\linewidth]{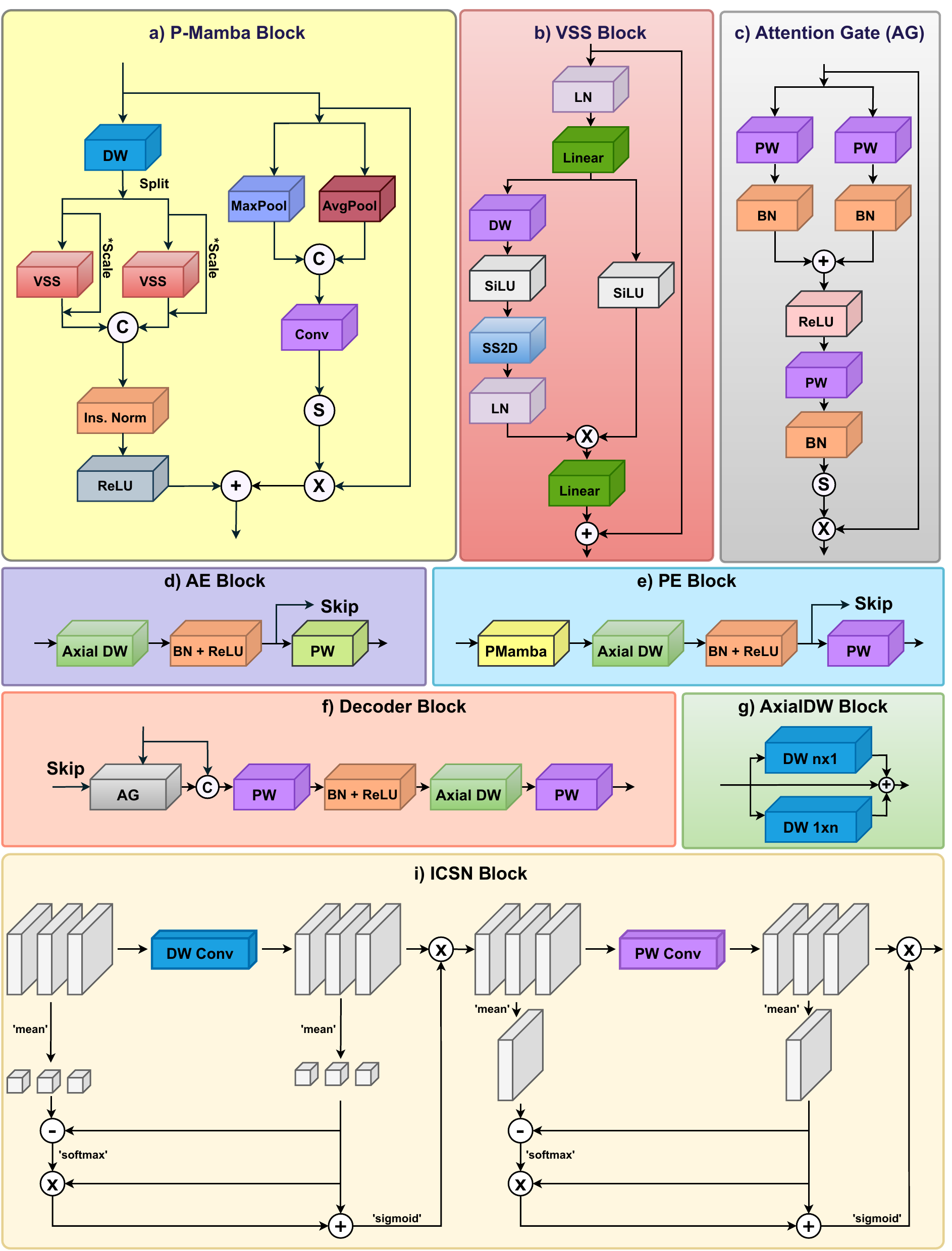}
\caption{The main components' architectures of the proposed MambaU-Lite model}
\label{fig:block}
\end{figure}

The proposed P-Mamba block, illustrated in Fig.\ref{fig:block}, is structured to improve the model's ability to learn diverse features.This is accomplished by processing the input feature maps through two distinct branches.

In the first branch, the input is passed through a Depthwise convolution layer with a 3x3 kernel to capture local features initially. To help reduce parameter count while maintaining stable performance, the input channels are split in half and fed into two VSS blocks, as shown in Fig.\ref{fig:block}. These blocks, introduced by Nguyen \textit{et al.} in AC-MambaSeg \cite{nguyen2024ac}, are designed to enable the model to learn multi-scale features effectively. The outputs of the two VSS blocks are concatenated to restore the original size and then normalized using Instance Normalization, followed by the ReLU activation function, which standardizes the output and enhances model stability.

In the second branch, the input is sequentially processed through AveragePooling and MaxPooling layers. Combining both pooling types allows the model to capture information at both global and detailed levels, enriching the feature representation. The outcomes of the pooling layers are concatenated and passed through a 3x3 Convolution layer to restore the channel count to its original size, which also helps the model focus on essential information. The output is then passed through a sigmoid function, which acts as an attention layer by emphasizing important features and suppressing irrelevant ones.

Finally, the outputs from the two branches are summed together, enabling the model to learn a broader variety of features. 



\subsection{The Encoder Block}
As described in Sec.\ref{The Proposed Model}, the encoder is composed of two main blocks: the AE Block and the PE Block, as shown in Fig.\ref{fig:block}d and Fig.\ref{fig:block}e. For the AE Block, the outputs are first processed through an AxialDW Convolution layer with a 7x7 kernel, introduced by Dinh \textit{et al.} \cite{dinh20231m}, subsequently undergoing Batch Normalization and activation via the ReLU function. Before proceeding to the Pointwise convolution layer to double the number of channels, a skip connection is extracted to avoid information loss and is used later in the Decoder. In the PE Block, the input is initially processed by the P-Mamba block, followed sequentially by an AxialDW Convolution layer with a 3x3 kernel, Batch Normalization, the ReLU activation function, and a Pointwise convolution layer. Similar to the AE Block, a skip connection is also extracted before Pointwise convolution layer to retain essential information for the decoding process.

\subsection{The Decoder Block}\label{The Decoder}
The overview of the Decoder block is presented in Fig.\ref{fig:block}f. Initially, the output from the previous Decoder layer is upsampled to match the size of the corresponding skip connection. It is then passed through the Attention Gate (AG) block, as shown in Fig.\ref{fig:block}c. The output of the AG block is concatenated with the upsampled feature maps from the previous Decoder layer. This concatenated output is then processed through a sequence of layers: a Pointwise convolution layer for dimensionality reduction, followed by Batch Normalization, ReLU activation, another Pointwise convolution, and finally an Axial Depthwise convolution with a 7x7 kernel. The combination of these layers enables the model to effectively extract meaningful features while minimizing the parameter count and computational overhead.

\subsection{The Skip Connection and Bottleneck Block}\label{The Skip connection and Bottleneck Block}
The skip connection and bottleneck components play crucial roles in the model, helping prevent information loss during processing. In the proposed model, we use the CBAM block introduced by Woo \textit{et al.} \cite{woo2018cbam} for the skip connections, while the bottleneck employs an Integrated Channel-Spatial Attention (ICSA) block.

The ICSA block consists of two consecutive Priority Channel Attention (PCA) blocks followed by a Priority Spatial Attention (PSA) block, as proposed by Le \textit{et al.} \cite{le2023attention}, which demonstrated outstanding performance in fish classification tasks. Specifically, the PCA block utilizes depthwise convolution to enhance feature extraction on each channel individually, while the PSA block applies pointwise convolution, improving feature maps across spatial regions. The use of the ICSA block in the bottleneck effective capture of high-level features effectively before passing them to the Decoder.
\section{Experiment}
\subsection{Dataset}
To assess the efficacy of the proposed method, we implement experiments on two skin lesion datasets:  ISIC 2018 and PH2. The ISIC 2018 dataset comprises 2,594 dermoscopic images along with segmentation masks. We divided this dataset into two parts: 2,334 images allocated for training and the remaining 260 images for testing. For the PH2 dataset, a smaller dataset with 200 images, we split it into two parts as well, with 170 images for training and 30 images for testing. Each image from both datasets was resized to 256x256 to facilitate the training process.
\subsection{Training and Evaluation Metric}\label{training}
We conducted experiments using the PyTorch framework, applying the Adam optimization strategy. The model underwent 300 epochs of training, with an initial learning rate of $1 \times 10^{-3}$, and the learning rate reduced by half if the Dice score did not improve after 10 consecutive epochs. For training, we used a composite loss function comprising Dice loss and Tversky loss. We set the hyperparameters for the Tversky loss as $\gamma_{1} = 0.7$ and $\gamma_{2} = 0.3$. The loss function formula is as follows:
\begin{equation}
L_{Dice}(g, p) = 1 - \frac{2 \sum_{i=1}^{n} y_i g_i}{\sum_{i=1}^{n} (g_i + p_i)}
\end{equation}
\begin{equation}
L_{Tversky}(g, p) = 1 - \frac{2 \sum_{i=1}^{n} g_i p_i}{\sum_{i=1}^{n} (g_i p_i) + \gamma_{1}\sum_{i=1}^{n} (g_i (1 - p_i)) + \gamma_{2}\sum_{i=1}^{n} ((1 - g_i) p_i)}
\end{equation}
\begin{equation}
L(g, p) = 0.5L_{Dice}(g, p) + 0.5L_{Tversky}(g, p)
\end{equation}
where \( g_i \in \{0, 1\} \) denotes the ground truth label, \( p_i \in (0, 1) \) refers to the predicted mask value for each pixel \( i \in \{1, 2, \ldots, N\} \), and \( N \) indicates the overall pixel count in the output segmentation mask.

To evaluate the model's performance, we utilize the two main metrics used in semantic segmentation: the Dice Similarity Coefficient (DSC) and Intersection over Union (IoU). These metrics help determine the similarity overlap between the predicted mask and the ground truth label, clearly demonstration of the effectiveness of the models.

\subsection{Results and Comparison}
To assess the proposed model's effectiveness, we compare it with previously proposed methods, encompassing U-Net \cite{ronneberger2015u}, Attention U-Net \cite{oktay2018attention}, UNeXt \cite{valanarasu2022unext}, DCSAU-Net \cite{xu2023dcsau}, and U-Lite \cite{dinh20231m}. These models were trained under conditions identical as the proposed model, and all implementations were sourced from the authors' open-source code repositories.
\begin{figure}[!h]
\centering
\includegraphics[width=\linewidth]{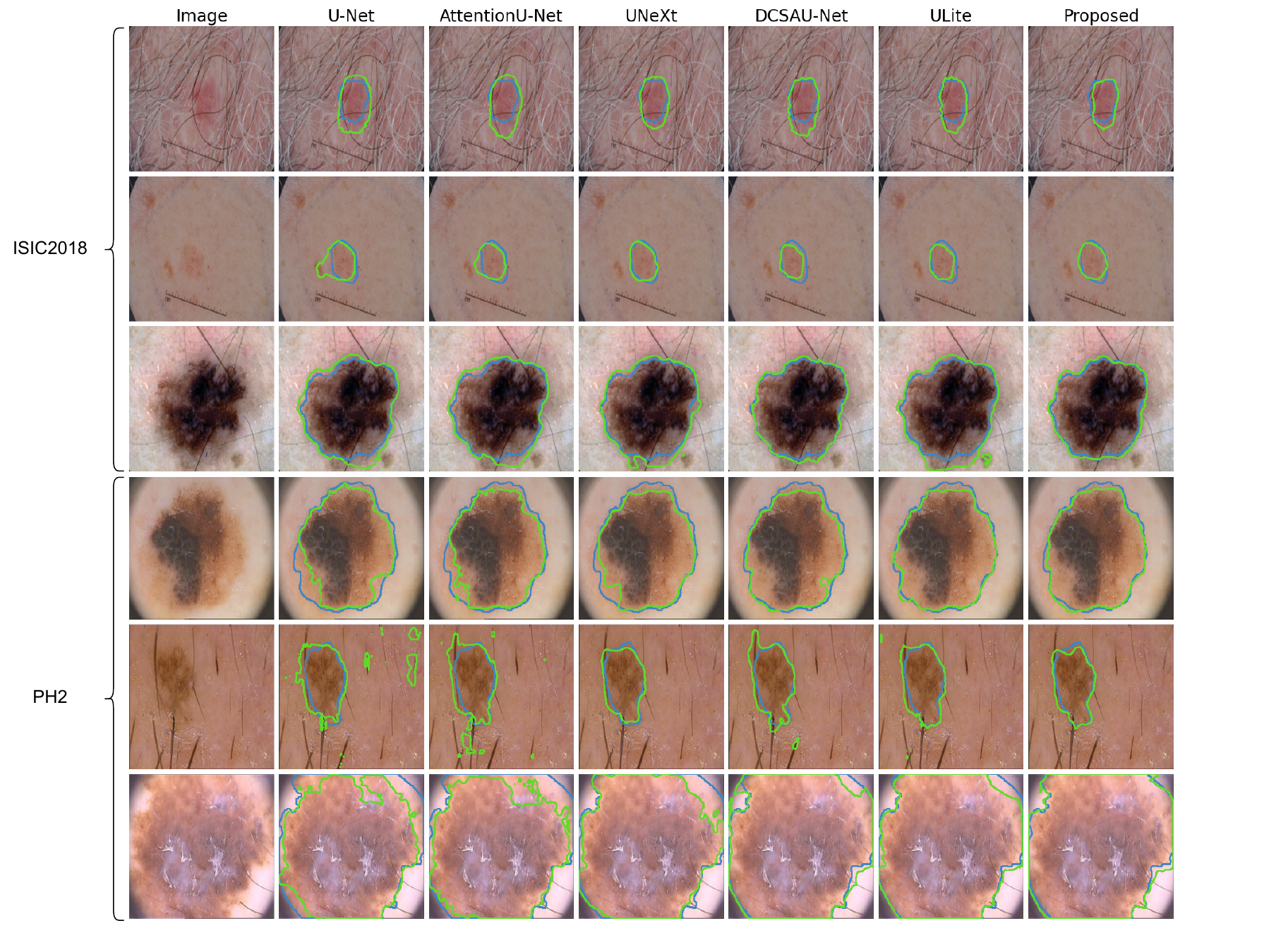}
\caption{Representative segmentation on the ISIC2018 and PH2 datasets.The ground truths are shown in blue, and the predictions are displayed in green.}
\label{fig:visual}
\end{figure}
The comparison results between MambaU-Lite and other models are conducted on the ISIC 2018 and PH2 datasets. Some visual segmentation results are illustrated in Fig.\ref{fig:visual}. As shown in this figure, the proposed MambaU-Lite model produces outputs more closely aligned with the original ground truth masks than other models, further affirming the accuracy and reliability of the segmentation model.

\begin{table}[!h]
\centering
\caption{\itshape Comparison on the ISIC2018 dataset.}
\label{isic}
\begin{tabular}{lccccc}
\toprule
\midrule
Methods & Params & FLOPS & Memory size & DSC & IoU \\ 
\midrule
U-Net \cite{ronneberger2015u} & 31.04M & 48.23G & 124.15MB & 0.8916 & 0.8176  \\
Attention U-Net \cite{oktay2018attention} & 34.88M & 66.54G & 139.51MB & 0.8965 & 0.8243 \\
UNeXt \cite{valanarasu2022unext} & 1.47M & \textbf{0.58G} & 5.89MB & 0.8983 & 0.8299 \\
DCSAU-Net \cite{xu2023dcsau} & 2.60M & 6.72G & 10.40MB &0.8929 & 0.8254 \\
U-Lite \cite{dinh20231m} & 0.88M & 0.69G & 3.51MB & 0.9032 & 0.8340\\
Proposed MambaU-Lite & \textbf{0.42M} & 1.25G &\textbf{1.67MB}& \textbf{0.9057} & \textbf{0.8361} \\
\midrule
\bottomrule
\end{tabular}
\end{table}

\begin{table}[!h]
\centering
\caption{\itshape Comparison on the PH2 dataset.}
\label{ph2}
\begin{tabular}{lccccc}
\toprule
\midrule
Methods & Params & FLOPS & Memory size & DSC & IoU \\ 
\midrule
U-Net \cite{ronneberger2015u} & 31.04M & 48.23G & 124.15MB & 0.9322 & 0.8775  \\
Attention U-Net \cite{oktay2018attention} & 34.88M & 66.54G & 139.51MB & 0.9287 & 0.8703 \\
UNeXt \cite{valanarasu2022unext} & 1.47M & \textbf{0.58G} & 5.89MB & 0.9409 & 0.8922 \\
DCSAU-Net \cite{xu2023dcsau} & 2.60M & 6.72G & 10.40MB &0.9416 & 0.8926 \\
U-Lite \cite{dinh20231m} & 0.88M & 0.69G & 3.51MB & 0.9483 & 0.9036\\
Proposed MambaU-Lite & \textbf{0.42M} & 1.25G &\textbf{1.67MB}& \textbf{0.9572} & \textbf{0.9189} \\
\midrule
\bottomrule
\end{tabular}

\end{table}
Quantitative comparison on the ISIC2018 in Table \ref{isic} shows that the proposed MambaU-Lite achieves superior performance over other models, with a DSC of 0.9057 and an IoU of 0.8361. The second-best performing model is Ulite, with a DSC of 0.9032 and an IoU of 0.8340. Although U-Lite has a lower FLOPS of 0.69G compared to the proposed model, it has significantly higher parameters and memory size, with 0.88M parameters and 3.51MB of memory, which is nearly twice as large as MambaU-Lite’s 0.42M parameters and only 1.67MB memory size.
The effectiveness of our model on a small dataset, PH2 is displayed in Table \ref{ph2}. It can be observed that MambaU-Lite model outperforms the other models, achieving a DSC of 0.9572 and an IoU of 0.9189. Additionally, our model has the lowest parameter count and memory size. Although UNeXt has a lower computational cost than the proposed model, its performance is comparatively lower, with a DSC of only 0.9409, which is significantly less than that of MambaU-Lite.


\section{Conclusion}
In this study, we introduced the lightweight MambaU-Lite model for the skin lesion segmentation, designed to minimize the number of parameters, computation cost, and memory usage. We proposed the P-Mamba block, integrated into an innovative architecture that combines the strengths of Mamba and CNNs to effectively capture both high-level and fine-grained features. While our model has shown promising results on skin lesion datasets, future work will aim to further optimize and generalize the model for a range of medical imaging tasks, enhancing its adaptability and making it well-suited for deployment in medical devices.
\section*{Acknowledgement}
This research is funded by Vietnam National Foundation for Science and Technology Development (NAFOSTED) under grant number 102.05-2021.34.
\bibliographystyle{IEEEtran}

\end{document}